%% file: orienteerijcai.tex
\documentclass{article}
\usepackage{ijcai21}

\usepackage{booktabs} 
\usepackage{amsmath}
\usepackage{hyperref}
\usepackage{verbatim}
\usepackage{graphicx}
\usepackage{multirow}
\usepackage{amsfonts}
\usepackage{amssymb}
\usepackage{mathrsfs}
\usepackage{array}
\usepackage{subfigure}
\usepackage{multicol}
\usepackage{color}
\usepackage{enumitem}
\usepackage{graphicx}
\usepackage{epsfig}
\usepackage{epsfig}
\usepackage{natbib}
\usepackage{xspace}
\usepackage{booktabs}
\usepackage{url}
\usepackage{algorithm}
\usepackage{algorithmic}
\usepackage{balance}

\newcommand{\secref}[1]{Section~\ref{#1}} 
\newcommand{\figref}[1]{Figure~\ref{#1}} 
\newcommand{\tbref}[1]{Table~\ref{#1}} 
\newcommand{\vpara}[1]{\vspace{0.1in}\noindent\textbf{#1 }}

\DeclareMathOperator*{\argmax}{\arg\!\max}

\newcommand{\hide}[1]{} 
\newcommand{\reminder}[1]{}
\title{Boosting Graph Search with Attention Network for Solving the General Orienteering Problem}
\author{
    
    \affiliations
    
    \emails
    
}
\author{
	Jing Xu
	\affiliations
	State Grid Corporation of China
	\emails
	xujing1237@163.com
}
\author{
Jintao Su
\affiliations
Zhejiang University
\emails
 jtsu@zju.edu.cn
 
 Tao Xiao
 \affiliations
 State Grid Corporation of China
 \emails
 xjtuxt@gmail.com
}
\author{
	Zongtao Liu$^1$\and
	Jing Xu$^2$\and
	Jintao Su$^1$\and
	Tao Xiao$^2$\And
	Yang Yang$^1$\\
	\affiliations
	$^1$Zhejiang University\\
	$^2$State Grid Corporation of China\\
	\emails
	tomstream@zju.du.cn,
	xujing1237@163.com,
	jtsu@zju.edu.cn, 
	xjtuxt@gmail.com,
	yangya@zju.edu.cn
}

\begin{document}

\maketitle
\input{abstract.tex}
\input{intro.tex}
\input{related.tex}
\input{setup.tex}

\input{model.tex}
\input{exp.tex}

\input{conclusion.tex}

\clearpage

\bibliographystyle{named}
\bibliography{reference}

\end{document}

%% file: abstract.tex
\begin{abstract}
Recently, several studies have explored the use of neural network to solve different routing problems, which is an auspicious direction. These studies usually design an encoder-decoder based framework that uses encoder embeddings of nodes and the problem-specific context to produce node sequence(path), and further optimize the produced result on top by beam search. However, existing models can only support node coordinates as input, ignore the self-referential property of the studied routing problems, and lack the consideration about the low reliability  in the initial stage of node selection, thus are hard to be applied in real-world.

In this paper, we take the orienteering problem as an example to tackle these limitations. We propose a novel combination of a variant beam search algorithm and a learned heuristic for solving the general orienteering problem. We acquire the heuristic with an attention network that takes the distances among nodes as input, and learn it via a reinforcement learning framework. The empirical studies show that our method can surpass a wide range of baselines and achieve results close to the optimal or highly specialized approach. Also, our proposed framework can be easily applied to other routing problems. Our code is publicly available\footnote{https://anonymous.4open.science/repository/7cb20ede-b50e-4a9a-99a5-e3c3626bf1a1/}.

\end{abstract}

%% file: intro.tex
\section{Introduction}
\label{sec:intro}

\hide{
1. what is the orienteering problem
2. recent trend in neural field and their limitation
3. another trend.
	
}

The orienteering problem(OP) is an important routing problem that originates from the sports game of orienteering. In this game, each competitor starts at a specific control point, visits control points in the tour, and will arrive at a destination point (usually the same as the start point). Each control point has an associated prize, and the travel between control points involves a certain cost. The goal is to select a sequence of points such that the total prize is maximized within the cost constraint. This problem relates to several practical applications, such as resource delivery, tourist trip guide and single-ring design in telecommunication networks(\cite{golden1987orienteering,souffriau2008personalized,thomadsen2003quadratic}). 
\citet{golden1987orienteering} have shown the OP to be an NP-hard problem, which motivates vigorous research into the design of heuristic solvers and approximation algorithms.

A recent trend in the machine learning community for solving routing problems is using deep neural networks to learn heuristic algorithms. With the help of other meta-algorithm such as beam search, some of these methods can achieve near start-of-the-art performance on several tasks (\cite{li2018combinatorial,kool2018attention,vinyals2015pointer,khalil2017learning,nazari2018reinforcement}). These tasks include the traveling salesman problem(TSP) and the orienteering problem(OP), and the vehicle routing problem(VRP). These studies deem solving such problems as a sequence generation
 process from the coordinates of nodes, and use recurrent neural networks(RNNs) or graph neural networks(GNNs) to build a bridge between node patterns and the policy. The OP belongs to the routing problem family and thus is fit for this methodology. 
 
 However, current learning-based algorithms have several limitations. 
Firstly, existing methods do not support direct input of travel cost information. They implicitly assume the travel cost, saying travel distance here, between each pair of points is their Euclidean distance, and take the coordinate of each position as network input to indirectly acquire the distance information. 
However, the distance metric used in real-world is not limited to the Euclidean distance, 
thus such kind of methods has limited practicability.
For the term convenience, we here define the OP which only supports coordinates as input as the \textit{coordinate orienteering problem}, and define the OP that supports direct input of costs among locations as the \textit{general orienteering problem}. 
Secondly, the routing problems considered in the previous works have self-referential property, but these works lack insight into it. For instance, in the orienteering problem, each step of point selection can be viewed as the initial step in another distinct orienteering problem, in which the competitor starts from the last selected points and will choose from the rest points.  Lacking consideration about such property in the existing studies might degrade the utilization of samples when training. 

Besides, most existing studies usually rely on a step-by-step beam search to acquire a better solution (\cite{kaempfer2018learning,vinyals2015pointer,nowak2017note}); however, it might not work well on the routing problem with relative large problem size. The step-by-step beam search at each step prunes non-promising partial solutions based on a heuristic function and stores a fix-sized set (also called \textit{beam}) of best alternatives. However, as the difficulty in the early stage of a routing problem is much more than that in the later stage, the estimated heuristic score (probability or action value) to perform beam search in the initial step might also be less credible. To overcome this shortcoming, the  stored partial solutions in the early stage of a search procedure should be more than those in the later stage (since the less credible heuristic scores might mislead the pruning). How to design such a search algorithm is an important issue.
 
 
 In this paper, we therefore propose a novel combination of a variant beam search algorithm and the learned heuristic for solving the general orienteering problem. We acquire the heuristic with an attention network that takes both node attribute (prize) and edge attribute (cost) as input, and learn it via a reinforcement learning framework.  Rather than formulating such routing tasks as sequence generation, we instead consider each step of point selection separately; that is,we view each state in the decision process as another different OP. This insight helps provide more direct reinforcement in each state and improves the utilization of training instances. The experimental result shows that our method can surpass a wide range of baselines and achieve results close to the optimal or highly specialized approach. 

\hide{
Accordingly, our contributions are as follows:
\begin{itemize} 
	\item We present a novel combination of an effective search algorithm and a learned heuristic for solving the general orienteering problem. This method can reduce the time and space complexity  in the search process and it is flexible in balancing the space, running time, and quality of plan.
	\item We designs an attention-based network that considers edge attributes to model routing problems. This is the first exploration to this type of problem that directly use edge attributes.
	\item By taking into account the self-referential property of such kind of problem, our method has better instances utilization compared with the previous encoder-decoder scheme.
	
\end{itemize}
}
\hide{
\vpara{Organization.} 
The remainder of this paper is organized as follows. 
We first formulate the problem and review some relevant research. We then give the necessary definitions and introduce the proposed approach, including the search algorithm, the attention model, and the learning algorithm. Subsequently, we present the experimental results and finally conclude the work. 
}

%% file: related.tex
\section{Related Work}
\vpara{Orienteering problem.}
Over the years, many challenging applications in logistics, tourism, and other fields were
modeled as OP. Meanwhile, quite a few studies for orienteering problems have been conducted. \citet{golden1987orienteering}  prove that the OP is NP-hard, i.e., no polynomial-time algorithm has been designed or is expected to be developed
to solve this problem to optimality. Thus the exact solution algorithms are very time consuming, and for practical applications, heuristics will be necessary.
Exact solution methods (using branch-and-bound and branch-and-cut ideas) have been presented by \cite{laporte1990selective,ramesh1992optimal}. Heuristic approaches, which use traditional operation research techniques, have been developed by \cite{tsiligirides1984heuristic,gendreau1998tabu,tang2005tabu}. 

\vpara{Applications of neural networks in routing problems.}
Using neural networks (NNs) for optimizing decisions in routing problems is a  long-standing direction, which can date back to the early work (\cite{hopfield1985neural,wang1995using}). These studies usually design an NN and learn a solution for each instance in an online fashion. Recent studies focus on using (D)NNs to learn about an entire class of problem instances. The critical point of these studies is to model permutation-invariant and variable-sized input appropriately.

Earlier researches leverage sequence-to-sequence models and the attention mechanism to produce path, and update network parameters via supervised or reinforcement learning (\cite{vinyals2015pointer,bello2016neural,nazari2018reinforcement}).
More recently, several studies leverage the power of GNNs that can handle variable-sized and order-independent input to tackle the routing problems (\citet{khalil2017learning,nowak2017note,kool2018attention}). However, these studies only focus on coordinate routing problem; thus the practicability is limited in real-word. Besides, current methods usually use a set-to-sequence/node-to-sequence model to perform routing, which ignores the self-referential property of many routing problems (including the OP).
\hide{
\vpara{Graph neural networks with node and edge information}
To handle general OP, how to learn with the distance among nodes in a complete graph becomes a vital problem. Most existing GNN methods focus on graphs with the only node attributes or further consider the edge weight, which might have limited applications since real-world graphs usually have relational attributes (i.e., from edges). Recently, several work has explored this direction(\cite{journals/corr/abs-1809-02709, conf/acl/CohnHB18, journals/corr/SchlichtkrullKB17,journals/corr/abs-1905-01436}). However, these methods can be time-consuming to tackle the complete graph considered in our paper as input, thus designing an effective and efficient way to model our problem is challenging.
\hide{\citet{journals/corr/SchlichtkrullKB17} introduce basis- and block-diagonal-decomposition to reduce the explosive parameters in the propagation on different kinds of edges(relations) in modeling relational data task. \citet{journals/corr/abs-1905-01436} improve the GNN architecture by making each layer has a node-update block and an edge-update block, and thus can iteratively update the edge-labels. } \reminder{address the importance of our methods}
}

%% file: setup.tex
\section{Problem Definition}
\label{sec:prob}
Formally, we define each orienteering problem instance $s$ as a tuple with six elements $\langle V, C, R, T, v_{start}, v_{end}\rangle$. More specifically, $V = \{v_1,...,v_M\}$ refers to the set of nodes (control points)  containing the start node $ v_{start}$, the prized nodes, and the end node $ v_{end}$, where $M$ denotes the number of nodes. $C$ refers to the cost map where $C_{v_i,v_j}$ represents the travel cost from $v_i$ to $v_j$. $R$ is the prize map where $R_{v}$ indicates the prize collected when $v$ is visited. 
Therefore,  the problem is to find a path from $v_{start}\in V$  to $v_{end} \in V$ such that the total cost of the path is less than the prescribed cost limit $T$, and the overall prize collected from the nodes visited is maximized.
The orienteering problem can also be formalized as a mixture integer problem, which can be referenced in \cite{gunawan2016orienteering}.

For notation convenience, here we introduce some operators or functions among these elements.  $cost(P)$, $prize(P)$, and $last(P)$ represent the total cost, the total prize, and the last selected point of the (partial) selected path $P$, respectively. $(P + v_i)$ represents a selected path obtained by adding a new node $v_i$ to the original path $P$. $(V-P)$ represents the node set that excludes the nodes in $P$ from $V$.

%% file: model.tex
\section{Method}
\label{sec:model}

\subsection{Cost-Level Beam Search with the Learned Heuristic}
Before introducing the search method proposed in this study, we first present an exact search method. We divide the maximal cost $T$ into $\lceil T/\tau \rceil$ intervals with the length of $\tau$. For each range, we maintain a queue to save all partial paths(solutions) with the total cost in that interval . Starting from interval $ t=0$, we iteratively retrieve the incomplete solution $P$ from the front of the queue and scan all the nodes that are unselected. For each node $v$ scanned, if $cost(P+v+v_{end})$ is no greater than the cost limit $T$,  we then add the updated partial path $(P+v)$ to the queue in interval $ (t +\lceil C_{last(P), v}/\tau \rceil) $. Finally, from all stored paths with the end node, the path with the highest total prize is the optimal path.

This method cannot finish computation in polynomial time. As the value of $t$ increases, the number of stored partial paths per window increases exponentially. To reduce the space and time occupied by the search, at each step of the iteration, it is practical to prune some partial paths based on a predefined heuristic score $f(P, s)$ with the input of the current partial solution $P$ and the problem instance $s$. To follow up this idea, we apply the idea of beam search and replace the queue maintained in each window $t$ by a priority queue sized $K$, which is used to save the paths with the K highest heuristic scores in each window.
We introduce a data-driven heuristic score function $f$ to estimate the total prize that can be reached along the partial path $P$: 
\begin{equation}
f(P, s) = prize(P) + e(s';\theta) \label{eq:f}
\end{equation} 
where $f$ consists of two terms, in which the term $prize(P)$ computes the total prize of the given partial path, and the term $e(s';\theta)$ is an estimation of the subsequent prize along the current path under $\pi$ till the end of the problem. $e(\cdot~;\theta)$ is an estimation function parameterized $\theta$ that computes the total potential prize given by an OP under the policy $\pi_\theta$ and $s'$ represents the subproblem $\langle V-P, C, S, T-cost(P), last(P), v_{end}\rangle$ of the original OP. 

A question that  naturally arises here is how to obtain a reliable function $e(\cdot~;\theta)$ to estimate the total prize of an orienteering problem. In the next section, we will present our solution with the attention-based neural networks learned by Q-learning.
\begin{algorithm}[!t]
	\caption{Cost-Level Beam Search}
	\label{alg:neural}
	\begin{algorithmic}[1]
		\STATE initialize a list of priority queues PQ[$\lceil T/\tau \rceil$]
		\STATE initialize the empty path $P_g$ and insert it to PQ[0]
		\FOR {t=0 to $\lceil T/\tau \rceil$}
		\WHILE {PQ[t] is not empty}
		\STATE pop partial path $P_c$ from PQ[t]
		\FOR {v in $(V-P_c)$}
		\STATE pq = PQ[$\lceil cost(P_c+v)/\tau\rceil$]
		\STATE \textbf{if}~ {$cost(P_c+v+v_{end})>= T$}~\textbf{then}~continue
		\STATE insert $(P_c+v)$ to pq
		\STATE \textbf{if}~$prize(P_c+v)>prize(P_g)$~\textbf{then}~$P_g=(P_c+p)$
		\IF{the size of pq is larger than K}
		\STATE pop $argmin_P f(P,s)$ from pq
		\ENDIF
		\ENDFOR
		\ENDWHILE
		\ENDFOR
		\RETURN path $P_g$
	\end{algorithmic}
\end{algorithm}
\hide{
The total prize function $prize(\cdot)$ can be used as an intuitive instantiation of the heuristic score function $f$. Compared with the optimal solution, such pruning rule leaves the paths with K highest total prizes in each window,  which significantly reduces the computational time and space.  However, it is still difficult to avoid some short-sighted trajectories. In other words, the stored paths with top K total prizes in the current window do not necessarily overlap with the actual optimal or the near-optimal track. To reduce the possible impact of this defect in the search process,
}

\subsection{Prize Estimation via Attention Network}
\citet{kool2018attention} propose a network architecture based on self-attention
 to model several coordinate routing problems, including TSP, OP, and several VRP variants. With the benefit of reinforcement learning, the learned heuristic achieves better performance over other learning-based heuristic methods. Following the idea of using the attention mechanism to model the node interactions with permutation-invariant property, we propose an attention-based network, named \textbf{A}ttention \textbf{N}etwork (AN). Our proposed structure is not limited in handling coordinate OP,  but can directly take all the node attributes (prize of each node), edge attributes (costs among nodes) and global attribute (the remaining cost of the agent) as input. 
\begin{figure}[t]
	\centering
	\includegraphics[width=0.45\textwidth]{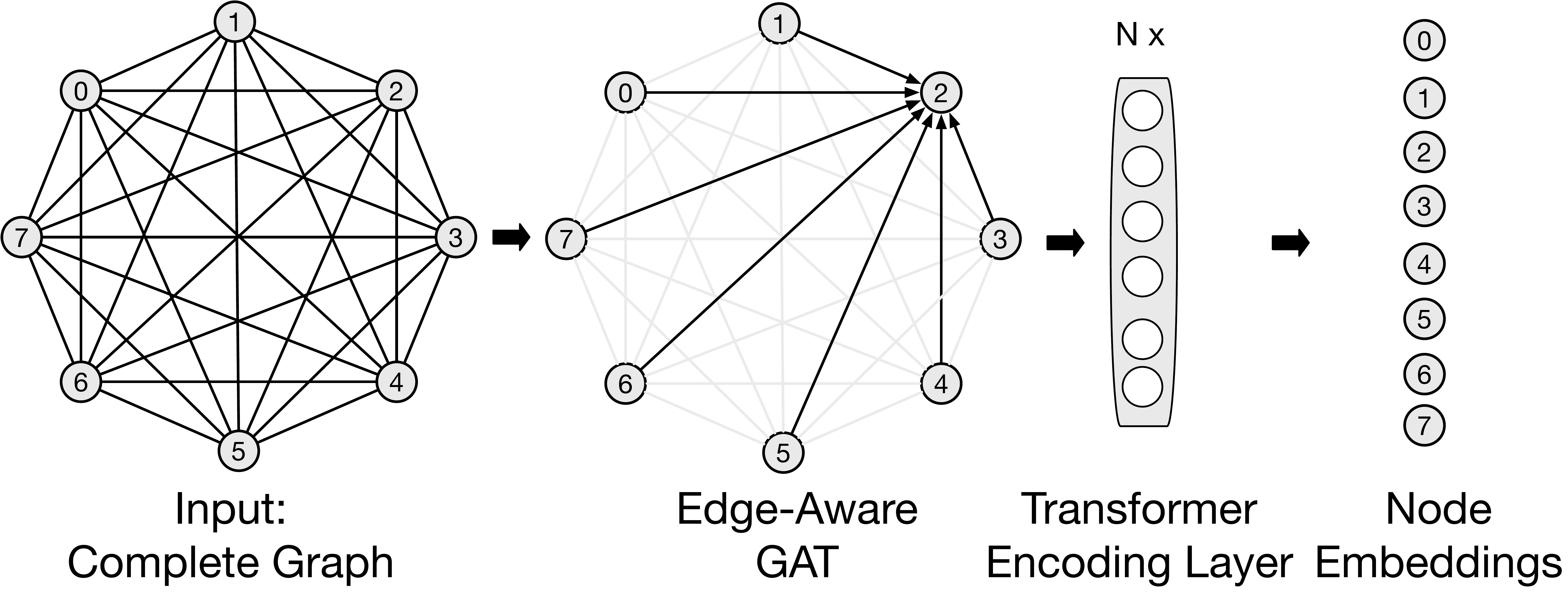}
	\caption{An illustration of our Attention Network.}
	\label{fig:model}
\end{figure} 

\vpara{Edge-Aware Graph Attention Networks Layer.}
The graph attention networks(GAT)\cite{velivckovic2017graph} utilize the attention mechanism to aggregate the attributes of node neighbors adaptively, and extract high-level representation feature vector of each node.
\hide{, which can be formulated as follows:
\begin{eqnarray}
\alpha_{vk} = \frac{\exp\left(\text{LeakyReLU}\left(\mathbf{a}^T[{\mathbf{W}}\mathbf{h}_v\|\mathbf{W}\mathbf{h}_k]\right)\right)}{\sum_{j\in\mathcal{N}_v} \exp\left(\text{LeakyReLU}\left(\mathbf{a}^T[{\bf W}\mathbf{h}_v\|{\bf W}\mathbf{h}_j]\right)\right)} \label{eqn:weight}\\
\mathbf{h}_{\mathcal{N}_v}' = \sigma\left(\sum_{k\in\mathcal{N}_v} \alpha_{vk} {\bf W}\mathbf{h}_k\right)\label{eqn:single}\\
MHA:
\mathbf{h}_{\mathcal{N}_v}' =  \sigma\left(\frac{1}{M}\sum_{m=1}^M \sum_{k\in\mathcal{N}_v}\alpha_{vk}^m{\bf W}^m\mathbf{h}_k\right) 
\end{eqnarray}
where $h_v\in \mathbb{R}^H$ denotes the representation of node $v$, $\sigma(\cdot)$ denotes the sigmoid function, $W$ and $U$ are the learnable parameter matrices, and $MHA$ is the abbreviation of multi-head average.
}
We modify the original structure of GAT; thus it can aggregate information from the connected edges, as well as the node neighbors and global information:
\begin{eqnarray}
\alpha_{vk} = \frac{\exp\left(\text{LeakyReLU}\left(\mathbf{a}^T\left[{\mathbf{W}}[\mathbf{x}_v\|\mathbf{x}_k\|\mathbf{u}_{vk}\|\mathbf{g}]\right]\right)\right)}{\sum_{j\in\mathcal{N}_v} \exp\left(\text{LeakyReLU}\left(\mathbf{a}^T\left[{\mathbf{W}}[\mathbf{x}_v\|\mathbf{x}_j\|\mathbf{u}_{vj}\|\mathbf{g}]\right]\right)\right)} \label{eqn:weight2}\\
\mathbf{h}_{\mathcal{N}_v} = \sigma\left(\sum_{k\in\mathcal{N}_v} \alpha_{vk}\left[{\mathbf{W}}[\mathbf{x}_v\|\mathbf{x}_k\|\mathbf{u}_{vk}\|\mathbf{g}]\right]\right)\label{eqn:single2}\\
MHA:
\mathbf{h}_{\mathcal{N}_v} =  \sigma\left(\frac{1}{M}\sum_{m=1}^M \sum_{k\in\mathcal{N}_v}\alpha_{vk}^m\left[{\mathbf{W}^m}[\mathbf{x}_v\|\mathbf{x}_k\|\mathbf{u}_{vk}\|\mathbf{g}]\right]\right) 
\end{eqnarray}
where $x_v$ denotes $v$'s node attribute (prize $R_v$), $u_{vw}$ denotes the edge attribute (travel cost $C_{v,w}$) between node $v$ and $w$, and $g$ indicates the problem-level feature (the maximal cost $T$).  
After aggregating information from connected edges and node neighbors, we obtain the intermediate representation of each node $H = \{h_{v_1}, h_{v_i}, ...,\\ h_{v_M}\}$, where $h_v$ is the representation of $v$. 

\vpara{Transformer Encoding Layer(TEL).}
Recently, \citet{kool2018attention} also demonstrates the effectiveness of the Transformer to model many routing problems dealing with coordinates of points. Following this idea, we use the encoding layer of the Tranformer in this study to extract the node features for the intermediate representation set $H = \{h_{v_1}, h_{v_2}, ..., h_{v_{M}}\}$. Since the resulting
node representations are invariant to the input order, we do not use a positional encoding.  
\hide{
Each Transformer encoding layer has two sub-layers, the first of which is a multi-head self-attention mechanism followed by a node-wise fully connected network. The residual connection is used\cite{he2016deep} around each of the two sub-layers, followed by layer normalization\cite{ba2016layer}.
To leverage these residual computations, the sub-layers in the encoder, as well as the embedding layers, produce outputs of dimension dmodel = 64. Because the use of the Transformer model is widespread, and the implementation used in this paper is almost the same as the implementation in \cite{vaswani2017attention}, we will omit the more detailed description of the structure of the Transformer encoding layer. 
}
 Feeding the immediate embeddings $H$ into $N$ stacked Tranformer encoding layers, we obtain the updated embeddings  $H' = \{h'_{v_1}, h'_{v_2}, ..., h'_{v_{M}}\}$.

\vpara{Linear Projection Layer.}
Finally, we use the linear combination of $v_i$'s embedding $h'_i$ to estimate the potential total prize $r_i$ when selecting $v_i$ in the next step.
\hide{\begin{equation}
q(s,v;\theta) = Uh'_v
\end{equation} 
}
where $U \in \mathbf{R}^{|H|}$ is a learnable weight vector.
Thus, we obtain the prize estimation $e(s;\theta) = max_{v\in V}q(s,v;\theta)$. Although it would be more convenient here to estimate the total prize of the given OP directly, we adopt this way for the sake of facilitating Q-learning to train the networks.

\subsection{Training via Q-learning}
\label{sec:qlearning} 
We use a fitted double Q-learning algorithm (\cite{van2016deep}) to obtain the potential future prize of each selection. 
\hide{Compared with fitting the optimal values generated by the exact approach like in \cite{vinyals2015pointer,li2018combinatorial}, the main advantage of this approach is that training with Q-learning can drastically save time spent, as the exact approach has very high time complexity. 
}
We formulate the states, actions, rewards, transitions, and policy in the reinforcement learning framework as follows:
\begin{itemize} 
	\item States: We define the state $s$ as the current subproblem of the original OP, i.e., $\langle V-P, C, S, T-cost(P), last(P), v_{end}\rangle$.
	
	\item Actions: The action $v$ is a node selection of $V$ that is in the part of the current state $s$. Both the definition of states and actions is applicable across node set with various size.
	
	\item Rewards:  We define the reward $r$ of each action $v$ as the prize collected when the node is visited.
	
	\item Transitions: A transition is a tuple $\langle s_t, v_t, r_{t+1}, s_{t+1}\rangle$ which represents the agent takes action $v_t$ in state $s_t$ and then observes the immediate reward $r_{t+1}$ and resulting state $s_{t+1}$.
	
	\item Policy: Given the current state $s$, we can compute the q-value $q(s,v;\theta)$ of each unselected node $v$ and apply a deterministic greedy policy $\pi(v|s) = \argmax_{v\in|V|}q(s,v;\theta)$.
\end{itemize}

We use double Q-learning (\cite{van2016deep}) to learn a greedy policy $\pi_\theta$ parametrized by the attention network and obtain the expected total prizes $q(s,v;\theta)$ of each action $v$ simultaneously.
The network parameters can be updated by minimizing the following loss function of the transitions picked up from the replayed memory.\hide{:  
\begin{eqnarray}
\mathop{\mathbb{E}}\limits_{s_t, a_t, s_{t+1}, r_{t+1}}\left[\left(y_t - q(s_t ,v_t;\theta)\right)^2\right] \label{eq:target}\\
y_t = r_{t+1} -q(s_{t+1} ,\argmax_v q(s_{t+1} ,v; \theta);\theta')
\end{eqnarray}
where $\theta'$ denotes parameters of the target Q-network updated periodically, and $\theta$ includes parameters of behavior Q-network outputting the action value for $\epsilon$-greedy policy, same as the algorithm described in \cite{van2016deep}. The motivation of adopting the double Q-learning instead of the vanilla Q-learning \cite{mnih2015human} is that the vanilla Q-learning suffers from overestimations of action values and the double Q-learning reduces such overestimations. 
}
More details can be referenced in supplementary part.

\hide{
\begin{algorithm}[!t]
	\caption{Double Q-learning for the Attention Network}
	\label{alg:neural}
	\begin{algorithmic}[1]
		\STATE Initialize replay memory D to capacity M
		\STATE Initialize attention networks with random weights  $\theta$ or pre-trained parameters
		\STATE Initialize the problem instance $s_t^i\gets new~Instance()$, $\forall i\in\{1,..,B\}$
		\FOR {count=1 to \textit{max\_iteration}}
		\FOR {t=1 to $T_{max}$}
		\STATE Choose node $v^i_t\gets argmax_{v_t^i}q(s_t^i, v_t^i; \theta)$ with probability $1-\epsilon$, otherwise choose an action $v^i$ randomly from the unselected nodes, $\forall i\in\{1,..,B\}$
		\STATE Obtain reward $r^i_{t+1}$ and next state $s_{t+1}^i$, $\forall i\in\{1,..,B\}$
		\STATE Store the transition $\langle s^i_t, v^i_t, r^{i}_t, s^i_{t+1}\rangle$ in $D$, $\forall i\in\{1,..,B\}$
		\STATE Update state $s_t^i\gets s_{t+1}^i$ if $s_{t+1}^i$ is not a state of termination, otherwise $s_t^i\gets new~Instance()$, $\forall i\in\{1,..,B\}$
		\ENDFOR
		\STATE Sample a batch of transitions $\{\langle s^i_t, v^i_t, r^{i}_t, s^i_{t+1}\rangle\}_{i=0}^{B'}$ from $D$
		\STATE Compute the target by \eqref{eq:target}
		\STATE Compute gradients of $\theta$ and apply updates 
		\ENDFOR
	\end{algorithmic}
\label{alg:qlearning}
\end{algorithm}
}

\subsection{Discussions}
\vpara{Comparison with the step-by-step beam search.}
The learned heuristic score function $f$ of \eqref{eq:f} can also be applied to the step-by-step beam search. However, as the routing in the early stage is much complicated than that in the later stage, the estimated heuristic score in the early stage might also be less reliable. In the initial stage of step-by-step beam search, the proportion of reliable selection could be very low and thus degrade the routing performance. Compared with the step-by-step beam search, the cost-level beam search store the early partial paths as many as possible.

\vpara{Running times.} As the computation of the estimated action values in each time window is parallelizable, the presented beam search algorithm can be significantly accelerated. The basic idea is to compute the action value on the GPU or TPU. 

\vpara{Extensibility to other routing problems.}
By simple modification, our framework can be easy to be extended to other routing problems. Here we take the classical TSP as an example. Similar to OP, we set the learned heuristic score function as $f(P, s) = -cost(P)-e(s', \theta)$, which represents the negative of the expected total cost of the whole tour, while the edge attribute is placed the same as those in the OP (the node and global attributes are not needed). The reward in each selection is defined as the negative of the increased cost. The cost constraint $T$ can be set as the total travel cost given by another simple heuristic method.

%% file: exp.tex
\section{Experimental Results}
\label{sec:exp}
\begin{table*}[ht!]
	\label{tab:results_problems}
	\centering
	\caption{Performance comparison. The gap \% is w.r.t. the best value across all methods.}
	\renewcommand{\arraystretch}{0.8}
	\begin{tabular}{l|l|rrr|rrr|rrr}
		\toprule
		& &  \multicolumn{3}{c|}{$n = 20 ,  T=2$} & \multicolumn{3}{c|}{$n = 50 ,  T=3$} & \multicolumn{3}{c}{$n = 100 ,  T=4$} \\
		Prize & Method &  \multicolumn{1}{c}{Obj.} & \multicolumn{1}{c}{Gap} & \multicolumn{1}{c|}{Time} & \multicolumn{1}{c}{Obj.} & \multicolumn{1}{c}{Gap} & \multicolumn{1}{c|}{Time} & \multicolumn{1}{c}{Obj.} & \multicolumn{1}{c}{Gap} & \multicolumn{1}{c}{Time} \\
		\midrule
		\multirow{14}{*}{{Uniform}}
		&  Gurobi  &  $\textbf{5.85}$ & $0.00 \%$ & (1s) & \multicolumn{3}{c|}{-} & \multicolumn{3}{c}{-} \\
		&  Compass  &  $5.84$ & $0.17 \%$ & (0s) & $\textbf{16.46}$ & $0.00 \%$ & (2s) & $\textbf{33.30}$ & $0.00 \%$ & (6s) \\
		\cmidrule{2-11}
		&  Random  &  $2.13$ & $63.59 \%$ & (0s) & $3.33$ & $79.77 \%$ & (0s) & $4.37$ & $86.88 \%$ & (0s) \\
		&  Tsili (greedy)  &  $4.85$ & $17.09 \%$ & (0s) & $12.46$ & $22.94 \%$ & (0s) & $25.48$ & $23.48 \%$ & (0s) \\
		&  AN-dqn (greedy)  &  ${5.15}$ & ${11.97 \%}$ & (0s) & ${13.91}$ & ${15.49 \%}$ & (0s) & ${26.51}$ & ${20.39 \%}$ & (1s) \\
		&  AN-a2c (greedy)  &  ${4.97}$ & ${15.04 \%}$ & (0s) & ${14.13}$ & ${14.16 \%}$ & (0s) & ${24.31}$ & ${27.00 \%}$ & (1s) \\
		&  AN-at (greedy)  &  ${3.98}$ & ${31.97 \%}$ & (0s) & ${8.22}$ & ${50.06\%}$ & (0s) & ${9.75}$ & ${70.72 \%}$ & (1s) \\
		&  AN-pn (greedy)  &  ${4.46}$ & ${23.76 \%}$ & (0s) & ${11.10}$ & ${32.56 \%}$ & (1s) & ${10.06}$ & ${69.79 \%}$ & (2s) \\
		\cmidrule{2-11}
		&  Tsili (beam)  &  $5.54$ & $5.30 \%$ & (0s) & $13.54$ & $15.63 \%$ & (2s) & $25.91$ & $22.19 \%$ & (9s) \\
		&  AN-dqn (beam)  &  ${5.74}$ & ${18.80 \%}$ & (1s) & ${15.76}$ & ${4.25 \%}$ & (10s) & ${30.61}$ & ${8.07 \%}$ & (40s) \\
		&  AN-a2c (beam)  &  ${5.57}$ & ${4.79 \%}$ & (1s) & ${15.13}$ & ${8.08 \%}$ & (5s) & ${27.40}$ & ${17.72 \%}$ & (23s) \\
		&  AN-at (beam)  &  ${4.45}$ & ${23.93 \%}$ & (1s) & ${9.45}$ & ${16.46 \%}$ & (4s) & ${10.55}$ & ${68.32 \%}$ & (20s) \\
		&  AN-pn (beam)  &  ${4.84}$ & ${17.26 \%}$ & (1s) & ${13.75}$ & ${42.59 \%}$ & (4s) & ${12.08}$ & ${63.72 \%}$ & (24s) \\
		
		\cmidrule{2-11}
		&  CS (20, 0.05)  &  ${5.82}$ & ${0.51 \%}$ & (1s) & ${16.13}$ & ${2.00 \%}$ & (7s) & ${30.93}$ & ${7.12 \%}$ & (40s) \\
		\midrule
		\multirow{14}{*}{{Distance}}
		&  Gurobi  &  $\textbf{5.39}$ & $0.00 \%$ & (4s) & \multicolumn{3}{c|}{-} & \multicolumn{3}{c}{-} \\
		&  Compass  &  $5.37$ & $0.37 \%$ & (0s) & $\textbf{16.17}$ & $0.00 \%$ & (3s) & $\textbf{33.19}$ & $0.00 \%$ & (8s) \\
		\cmidrule{2-11}
		&  Random  &  $1.89$ & $64.94 \%$ & (0s) & $2.90$ & $82.07 \%$ & (0s) & $3.81$ & $88.52 \%$ & (0s) \\
		&  Tsili (greedy)  &  $4.08$ & $24.30 \%$ & (0s) & $12.46$ & $22.94 \%$ & (0s) & $25.69$ & $22.60 \%$ & (0s) \\
		&  AN-dqn (greedy)  &  ${4.77}$ & ${11.50 \%}$ & (0s) & ${13.70}$ & ${15.28 \%}$ & (0s) & ${26.60}$ & ${19.86 \%}$ & (1s) \\
		&  AN-a2c (greedy)  &  ${4.57}$ & ${15.21 \%}$ & (0s) & ${13.70}$ & ${15.28 \%}$ & (0s) & ${22.32}$ & ${32.75 \%}$ & (1s) \\
		&  AN-at (greedy)  &  ${3.62}$ & ${32.84 \%}$ & (0s) & ${9.29}$ & ${42.55 \%}$ & (1s) & ${10.52}$ & ${68.30 \%}$ & (2s) \\
		&  AN-pn (greedy)  &  ${3.75}$ & ${30.43 \%}$ & (0s) & ${11.37}$ & ${29.68\%}$ & (1s) & ${10.77}$ & ${67.55 \%}$ & (2s) \\
		\cmidrule{2-11}
		&  Tsili (beam)  &  $5.26$ & $1.68 \%$ & (0s) & $15.50$ & $14.50 \%$ & (2s) & $30.53$ & $8.04 \%$ & (10s) \\
		&  AN-dqn (beam)  &  ${5.27}$ & ${11.5 \%}$ & (1s) & ${15.82}$ & ${4.14 \%}$ & (9s) & ${30.28}$ & ${8.77 \%}$ & (35s) \\
		&  AN-a2c (beam)  &  ${5.02}$ & ${6.86 \%}$ & (1s) & ${15.13}$ & ${6.43 \%}$ & (6s) & ${27.83}$ & ${16.15 \%}$ & (24s) \\
		&  AN-at (beam)  &  ${4.42}$ & ${18.00 \%}$ & (1s) & ${10.68}$ & ${33.95 \%}$ & (4s) & ${11.10}$ & ${66.56\%}$ & (21s) \\
		&  AN-pn (beam)  &  ${4.24}$ & ${21.34 \%}$ & (1s) & ${12.15}$ & ${24.86\%}$ & (4s) & ${12.48}$ & ${62.40 \%}$ & (17s) \\	
		\cmidrule{2-11}
		&  CS (20, 0.05)  &  ${5.35}$ & ${0.74 \%}$ & (1s) & ${15.99}$ & ${1.11 \%}$ & (6s) & ${31.16}$ & ${6.12\%}$ & (42s) \\
		\bottomrule
	\end{tabular}
	\label{tb:performance}
\end{table*}
\subsection{Dataset Creation}
To evaluate our proposed method against other algorithms, we generate problem instances with different settings. We sample each node $v$'s coordinates $x_v$ uniformly at random in the unit square and then compute the travel cost from node $v$ to node $w$ by their Euclidean distance $dis(v, w)$, where $dis(v, w)$ denotes the Euclidean distance between node $v$ and $w$. We set the distance metrics as Euclidean distance because although many OP algorithms can handle the OP with different distance metrics, most released implementations can only support planar coordinates as input for the sake of convenience. One more to mention is that selecting the Euclidean distance metric does not mean that we will compare the performance of our approach with those specialized methods designed for coordinate OP like \cite{khalil2017learning, kool2018attention}, because such comparison is unfair.

For the prize distribution, we consider three different settings described by \cite{fischetti1998solving,kool2018attention}. Besides, we normalize the prizes to make the normalized value fall between 0 and 1.
\hide{
\vpara{Constant.} $r_v=1$. Each node shares a fixed prize, so the agent aims to visit as many nodes as possible within the cost constraint.
}
\vpara{Uniform.} $r'_v\thicksim DiscreteUniform(1,100); r_v=r'_v/100$. The prize of each node is discretized uniformed.

\vpara{Distance.} $r_v = (1 + \lfloor99\cdot \frac{dis(x_{v_{start}}, x_v)}{max_{v\in V}dis(x_{v_{start}}, x_v)}\rfloor)/100$. Each node has a (discretized) prize that is proportional to the distance to the start node, which is a challenging task as the node with the largest prize are the furthest away from the start node\cite{fischetti1998solving}.

We choose the maximal travel cost $T$  as a value close to half of the average optimal length of the corresponding TSP tour. Setting approximately the half of the nodes can be visited results in the most challenging problem instances (\cite{vansteenwegen2011orienteering}).
Finally, we set the value of $T$ for the OP sized 20, 50, 100 as 2, 3, and 4, respectively.

\hide{
\subsection{Model Learning}
 For the hyperparameters of the network, we set the hidden size as 32, the number of multi-heads in both variant-GAT and Transformer as 8 and the number of stacked Transformer encoding layers as 4. For the hyperparameters of q-learning, we set the size of replay memory as 1024, and use Adam optimizer\cite{kingma2015adam} with the learning rate of $1\times10^3$ to update the model parameters. The number of training instances is 100K and the maximal iteration is set as 20K. \reminder{add hardware }
}
\subsection{Performance Comparison}
\label{sec:baselines}
\vpara{Comparable Methods.}We evaluate the performance of our method under the above-mentioned instance generation settings . Firstly, we compare our proposed method with an exact method \textit{Gurobi} (\cite{optimization2014inc}) and a state-of-the-art heuristic method \textit{Compass} (\cite{kobeaga2018efficient}).We further compare with several policy-based methods, which compute the node selection probability $p(v|s)$ or action value $q(s,v)$ in state $s$.These methods include (1)Random selection (\textit{Random}), (2)\textit{Tsili} (\citet{tsiligirides1984heuristic}), (3)policy learned by deep Q-learning algorithm (\textit{AN-dqn}), (4)policy learned by advantage actor-critic algorithm (\textit{AN-a2c}). We report their results based on two meta-algorithm: (1)greedy: the node with the highest probability or action value is selected in each state. (2) beam: we apply a step-by-step beam search for some of the below methods to obtain the target path. The beam size is set as 100.

\hide{
\begin{itemize} 
	\item \textit{Gurobi:}  This baseline is an exact method solved by the Gurobi optimizer\cite{optimization2014inc}.
	\item \textit{Compass:}  This is a state-of-the-art Genetic Algorithm for solving the OP\cite{kobeaga2018efficient}.
\end{itemize}
}

\begin{figure}[h]
	\centering
	\subfigure[With/without MHA in GAT.\label{fig:gat}] {
		\includegraphics[width=0.14\textwidth]{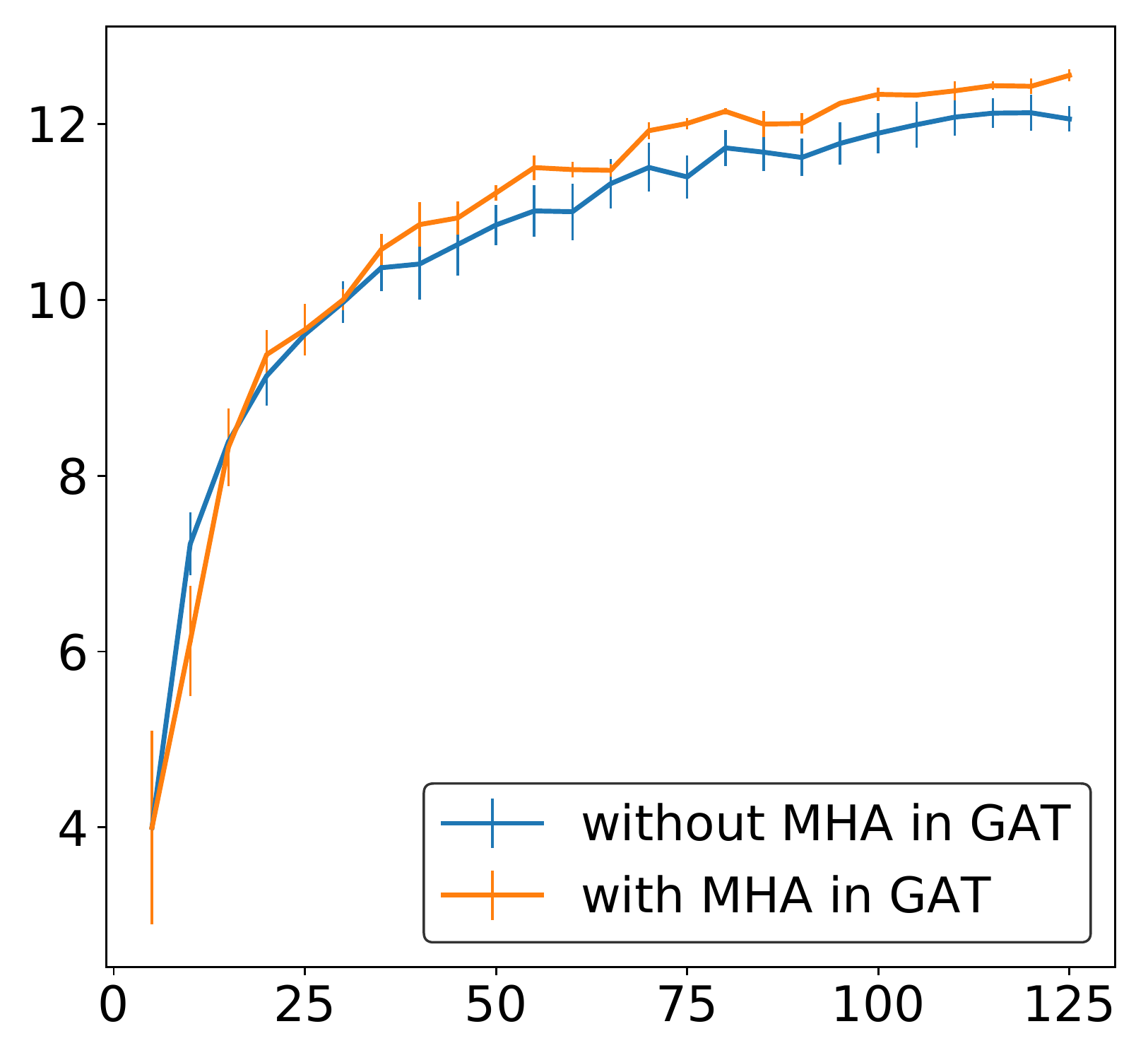}
	}
	\subfigure[Varing \#layer of TLE.\label{fig:layer}] {
		\includegraphics[width=0.14\textwidth]{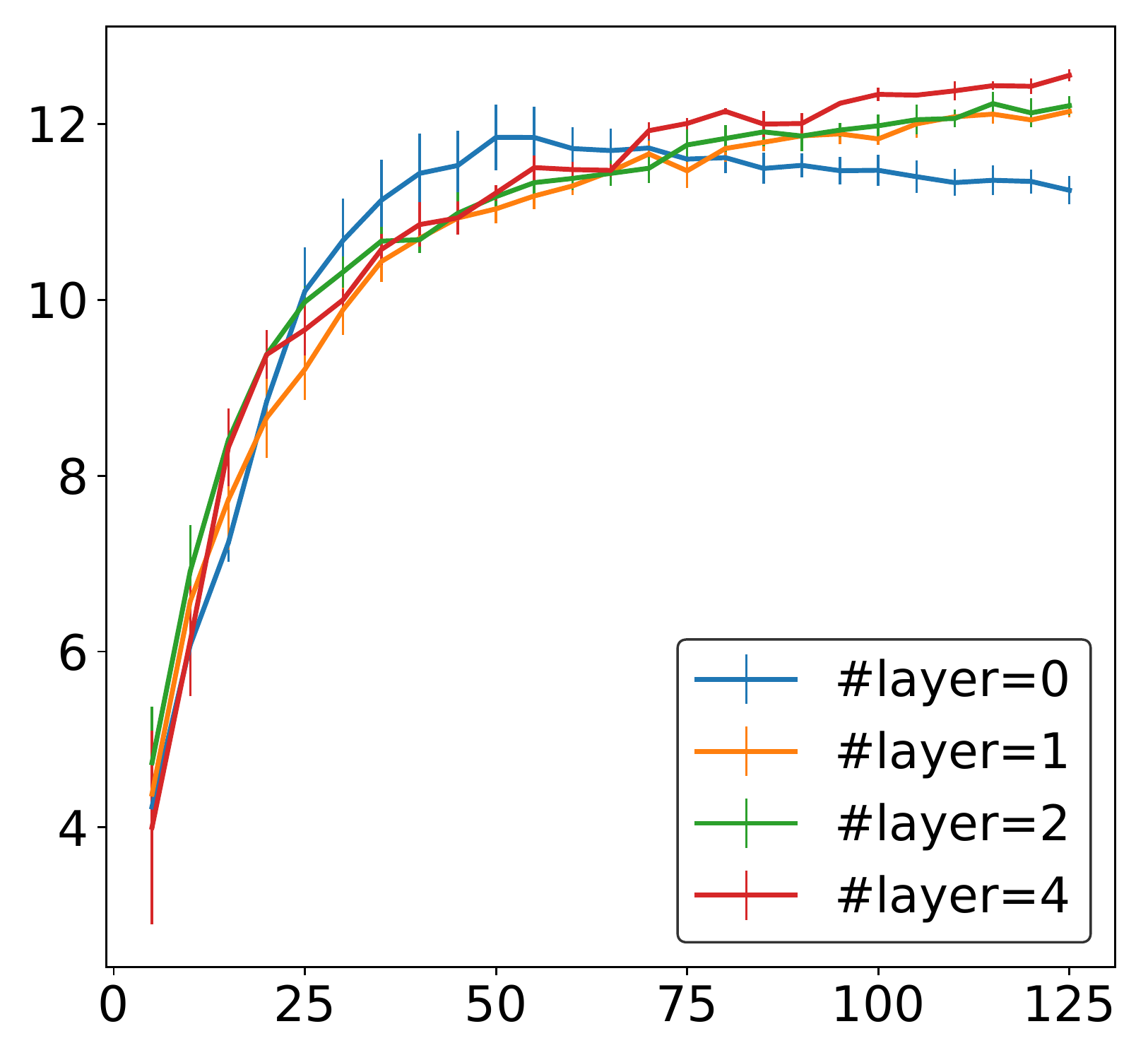}
	}
	\hfill
	\subfigure[Varing hidden size.\label{fig:hidden}] {
		\includegraphics[width=0.14\textwidth]{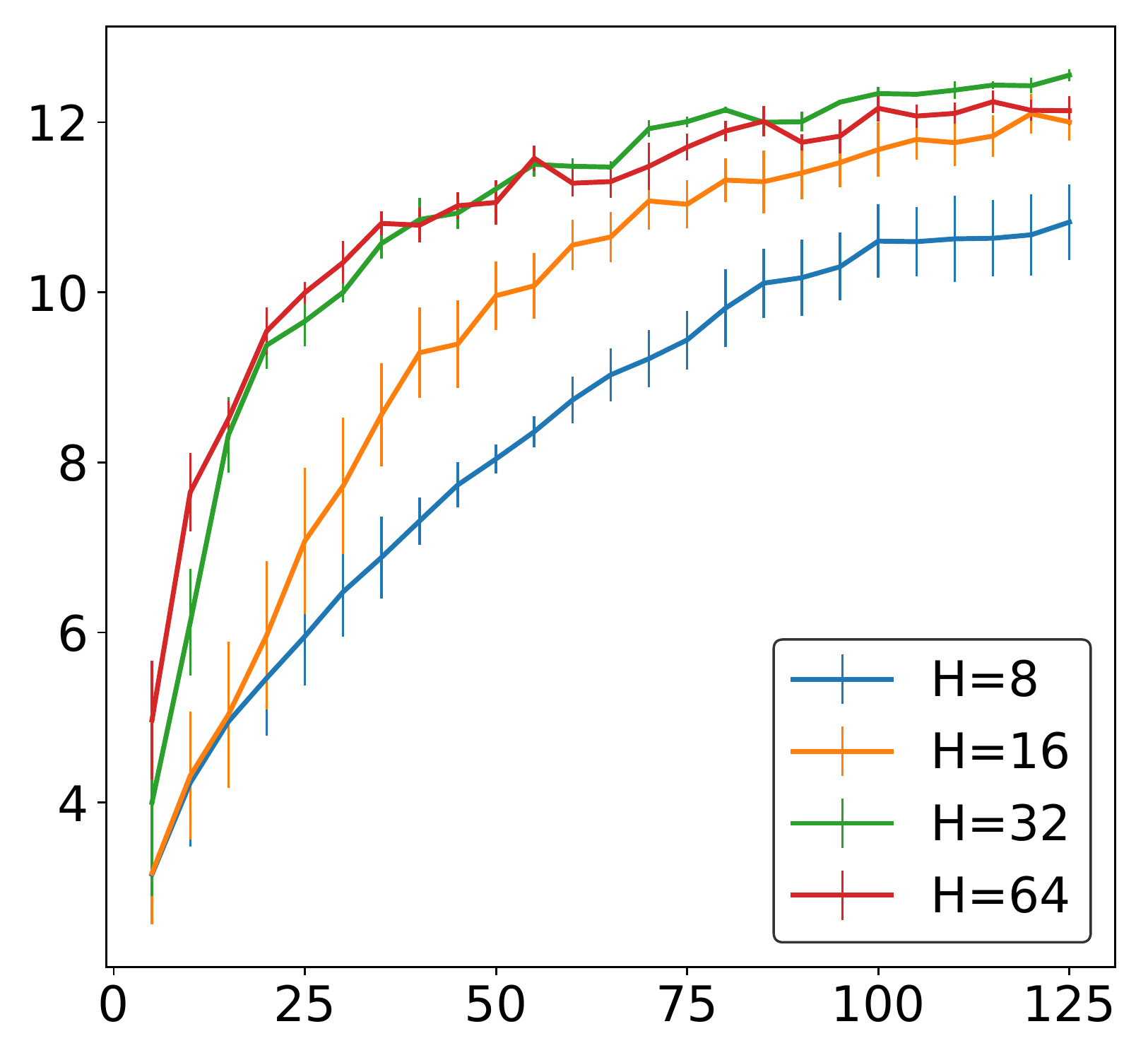}
	}
	\hfill
	\caption{Model parameter/structure analysis. The x-axis is the number of episodes. The y-axis denotes the average total prize on 200 test instances with greedy policy using the estimated action values from AN. This experiment is conducted five times, and we show the average performance and the corresponding standard error.}
	\label{fig:sensitive}
\end{figure}
In addition to the above methods as benchmark approaches, the comparison between $AN-a2c$, $AN-at$, and $AN-pn$ also serves to demonstrate the effectiveness of our network architecture compared with the encoder-decoder schemes used in the previous works. This is because $AN-a2c$ can be view as a policy gradient version of our proposed DQN framework proposed in \secref{sec:qlearning}, which considers each step of node selection as a distinct OP. In contract, $AN-at$ and $AN-pn$ view each selection is related to the previous ones (via RNN or self-attention mechanism).
Lastly, we report the performance of our cost-level beam search algorithm \textit{CS(K, $\tau$)}. $K$ and $\tau$ represent the beam size and cost interval, respectively.
\hide{
\begin{itemize} 
	\item \textit{CS-p (K, $\tau$):}  This approach refers to the cost-level beam search with $prize(\cdot)$ as the heuristic score function. $K$ and $\tau$ represent the beam size and cost interval, respectively.
	\item \textit{CS-l (K, $\tau$):}  This is our proposed method, which represents the cost-level search with the learned heuristic score function.
\end{itemize}
}

\vpara{Comparison Results.}
\tbref{tb:performance} reports the performance and average running time of each approach on 10K test instances. Presenting running time is not for the comparison between the efficiency of different methods. The results of different approaches might not be comparable due to the differences in their hardware usages (e.g., CPU or GPU), details of implementation (e.g., C++/Python with/without some optimizations), the settings of hyperparameters (e.g., beam size). We report the time spent mainly to show that in general, our method and implementation are time-affordable with the help of GPU acceleration.

We note that our search method with the learned heuristic ($CS$) surpasses all comparative methods except the exact approach \textit{Gurobi} and the specialized genetic algorithm \textit{Compass} (the gap is relatively small compared with those of other benchmarks), which demonstrates the effectiveness of our proposed methods. \textit{AN-a2c} is a competitive method concerning other baselines with attention network as a shared bottom. However, since the step-by-step beam search retains relative fewer partial paths at the start of the selection process, our search method significantly outperforms this variant. 

The comparison between $AN-dqn(beam)$ and $CS$ demonstrates the effectiveness of the cost-level beam search.

By comparison of the result of \textit{AN-a2c}, \textit{AN-at}, and \textit{AN-pn} in a greedy fashion or a beam search fashion, we can find which scheme of network design works better for the OP. We find that \textit{An-a2c} significantly outperforms the rest two approaches. This shows it is necessary to take into account the self-referential property of the OP and use a non-sequential network to model this problem.

\subsection{Parameter/Structure Analysis}
We conduct sensitive analysis in this section. We present the performance of AN-dqn (greedy policy using DQN) on 200 instances in the learning process. Each instance has 50 prized nodes, and the prize of each node is proportional to its distance to the start node (\textit{distance}).  We note that by introducing multi-head, the performance increases, which shows that the multi-head helps the GAT better aggregate edge information. Following the GAT, the Transformer encoding layer(TEL) further extracts more high-level representation for nodes. From \figref{fig:layer}, we observe that the performance improves as the number of layer increases. Lastly, we explore the relationship between the hidden size $\mathbb{R}^H$ and the quality of the learned model. 

\hide{
\subsection{Generalization Analysis}

To evaluate the capability of generalization of our method, we test the generalization performance on different problem sizes that trained for. We report results on 10K instances with the prize type as \textit{distance} (Cf. \tbref{tb:general}), because \textit{distance} is considered the hardest of these prize types. The performance drop of 2.5\% with the heuristic trained for the size of 20, and 0.99\% with the heuristic trained for the size of 50. The performance change is tiny and still outperforms most in \tbref{tb:performance}, which shows that our method has a good capability of generalization.

\begin{table}[ht!]
	\centering
	\renewcommand\arraystretch{1.2}
	\begin{tabular}{|l|c|c|c|}
		\hline
		\textbf{Problem size in training} &20 &50 &100 \\ \hline
		\textbf{Performance in testing} &30.38&30.85&31.16 \\ 
		\hline
	\end{tabular}
	
	\caption{Generalization Analysis. We list the performance of $CS-l$ with the heuristic trained on different problem size (20/50/100). The prize type is set as $distance$, while the testing problem size is set as 100.}\label{tb:exp:real}
	\label{tb:general}
\end{table}	
}
\subsection{Visualization}
\begin{figure}[h]
	\centering
	\subfigure[result of AN-dqn on Instance 1] {
		\includegraphics[width=0.21\textwidth]{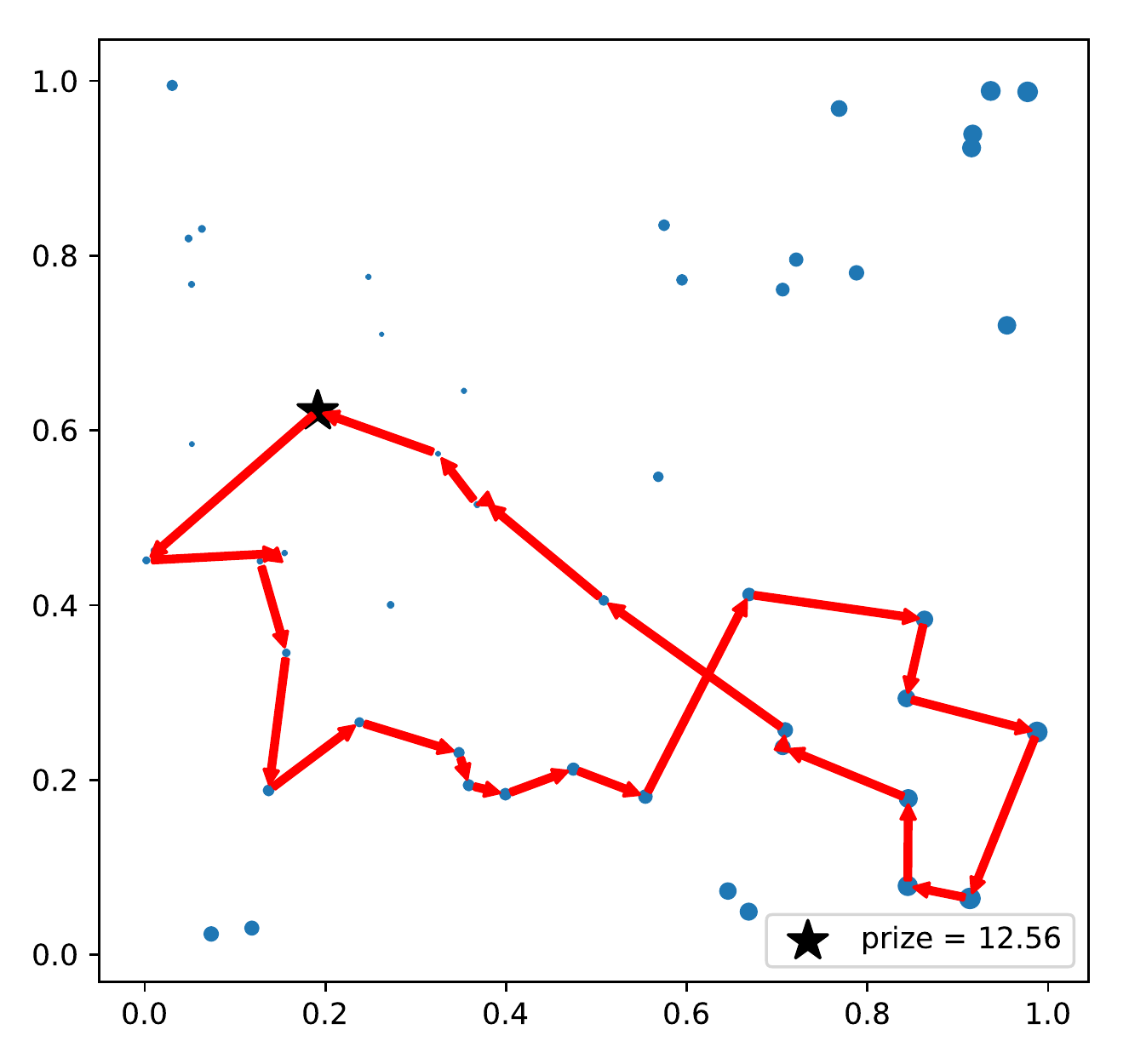}
	}
	\subfigure[result of AN-dqn on Instance 2] {
		\includegraphics[width=0.21\textwidth]{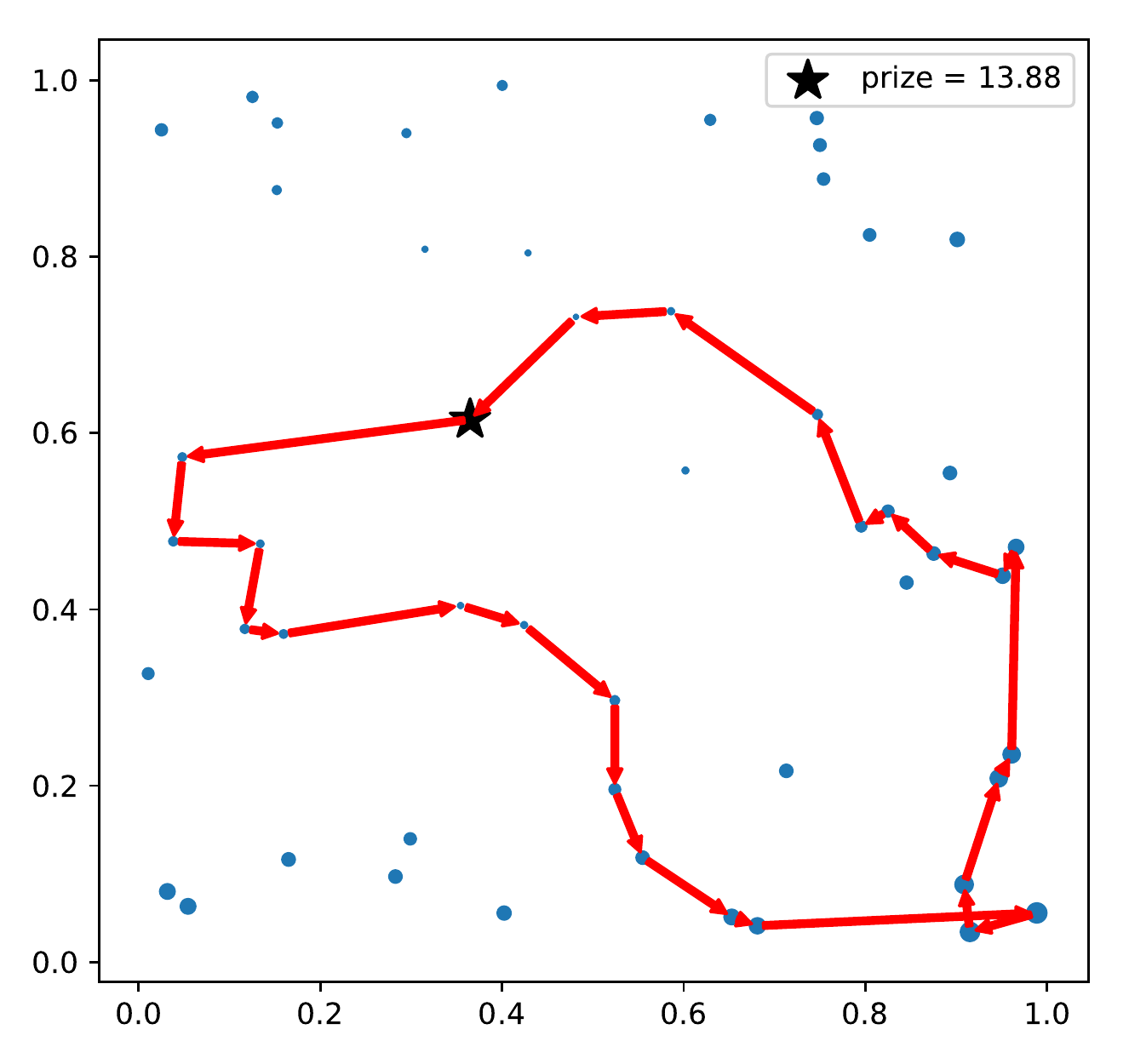}
	}
\hide{
	\subfigure[result of CS-l on Instance 1] {
		\includegraphics[width=0.18\textwidth]{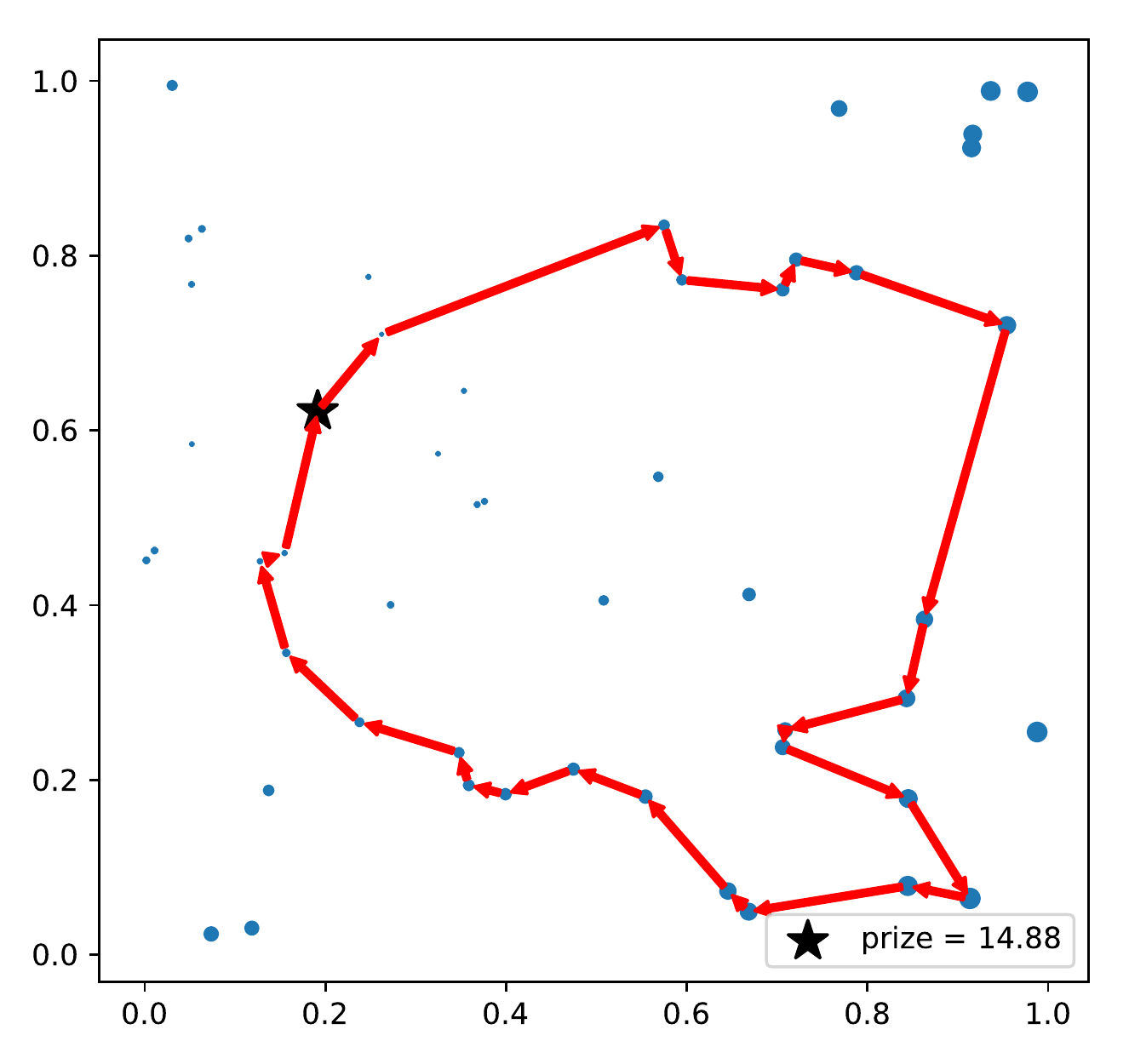}
	}
	\subfigure[result of CS-l on Instance 2] {
		\includegraphics[width=0.18\textwidth]{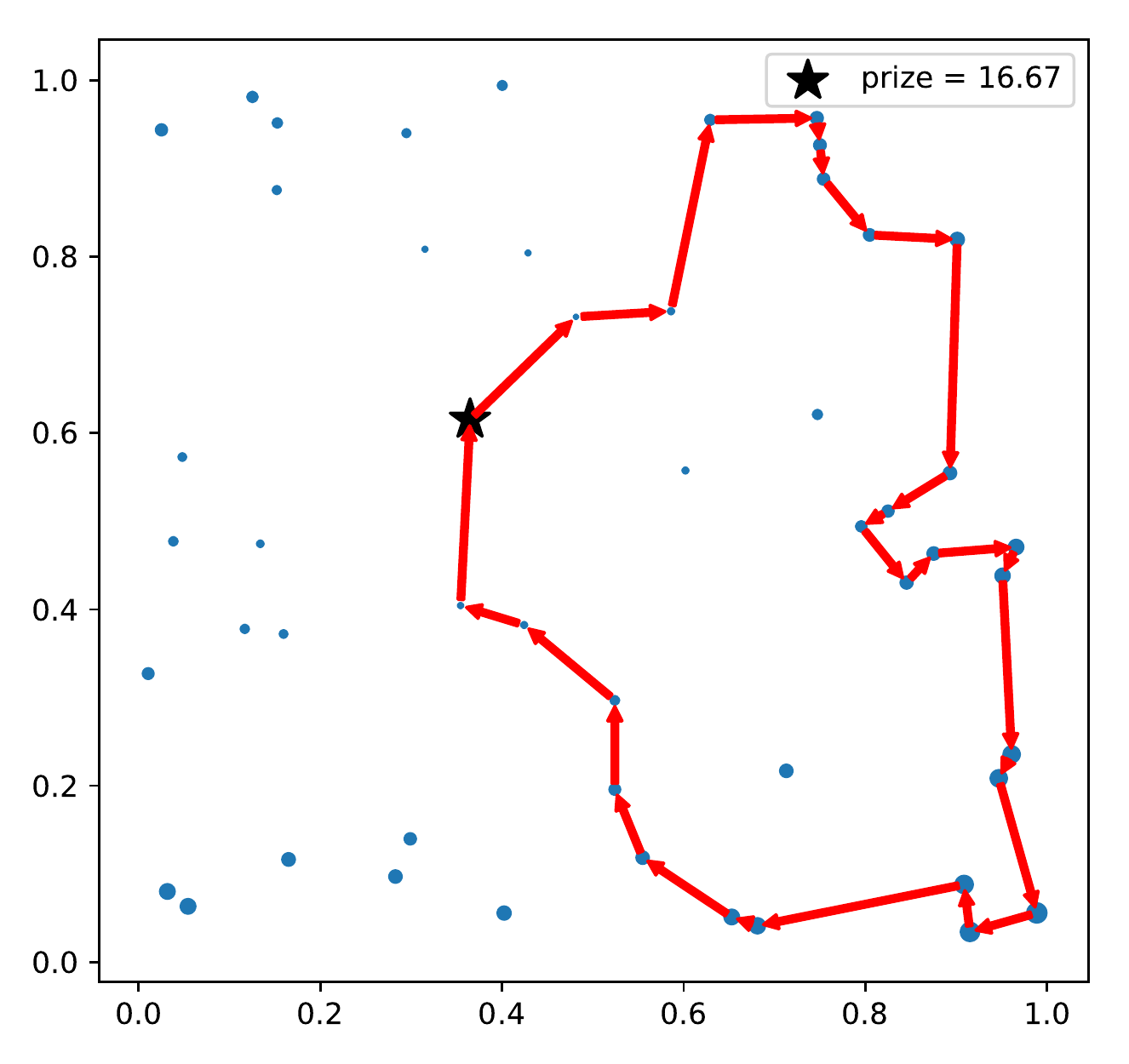}
	}
}
	\hfill
	\caption{Visualization of the output paths by AN-dqn, and CS. The star marker denotes the start/end node, and the circle marker represents the prized node, the area of which is proportional to its prize.}
	\label{fig:vis}
\end{figure}

Finally, we visualize the output paths generated by $AN-dqn$, $CS-p$, and $CS-l$ with the problem size of 50 and the prize type of \textit{distance} in \figref{fig:vis}. We observe that compared with the results of the greedy selection with action values ($AN-dqn$) or the cost-level beam search with a simple heuristic function ($CS-p$), the combination of our cost-level beam search and the learned heuristic ($CS-l$) can conduct more look-forwarding selection, thus have better performance.

%% file: conclusion.tex
\section{Conclusion and Future Work}
\label{sec:conclude}
In this paper, we solve the NP-hard problem of the general orienteering problem by conducting a novel combination of a cost-level beam search method and the learned heuristic score function. To estimate the potential prize of the current partial solution, we design an attention-based network. It first uses a variant GAT to aggregate both the edge attribute(travel cost) and the node attribute(prize), then applies stacked Transformer encoding layers to produce high-level node representations. Rather than modeling the OP in an encoder-decoder structure, we use the extracted node representations to directly estimate the target action values, which is a better scheme for the OP with the self-referential property. We further conduct sufficient experiments to show that our method surpasses most of the benchmark methods and get results close to those of the exact method and the highly optimized and specialized method. We also note that the performance of our method has been improved by introducing the deep neural network. Besides, with a simple modification, our proposed framework can also be straightforward to apply to other routing problems.

%% file: orienteerijcai.bbl
\begin{thebibliography}{}

\bibitem[\protect\citeauthoryear{Bello \bgroup \em et al.\egroup
  }{2016}]{bello2016neural}
Irwan Bello, Hieu Pham, Quoc~V Le, Mohammad Norouzi, and Samy Bengio.
\newblock Neural combinatorial optimization with reinforcement learning.
\newblock {\em arXiv preprint arXiv:1611.09940}, 2016.

\bibitem[\protect\citeauthoryear{Fischetti \bgroup \em et al.\egroup
  }{1998}]{fischetti1998solving}
Matteo Fischetti, Juan Jose~Salazar Gonzalez, and Paolo Toth.
\newblock Solving the orienteering problem through branch-and-cut.
\newblock {\em INFORMS Journal on Computing}, 10(2):133--148, 1998.

\bibitem[\protect\citeauthoryear{Gendreau \bgroup \em et al.\egroup
  }{1998}]{gendreau1998tabu}
Michel Gendreau, Gilbert Laporte, and Fr{\'e}d{\'e}ric Semet.
\newblock A tabu search heuristic for the undirected selective travelling
  salesman problem.
\newblock {\em European Journal of Operational Research}, 106(2-3):539--545,
  1998.

\bibitem[\protect\citeauthoryear{Golden \bgroup \em et al.\egroup
  }{1987}]{golden1987orienteering}
Bruce~L Golden, Larry Levy, and Rakesh Vohra.
\newblock The orienteering problem.
\newblock {\em Naval Research Logistics (NRL)}, 34(3):307--318, 1987.

\bibitem[\protect\citeauthoryear{Gunawan \bgroup \em et al.\egroup
  }{2016}]{gunawan2016orienteering}
Aldy Gunawan, Hoong~Chuin Lau, and Pieter Vansteenwegen.
\newblock Orienteering problem: A survey of recent variants, solution
  approaches and applications.
\newblock {\em European Journal of Operational Research}, 255(2):315--332,
  2016.

\bibitem[\protect\citeauthoryear{Hopfield and Tank}{1985}]{hopfield1985neural}
John~J Hopfield and David~W Tank.
\newblock “neural” computation of decisions in optimization problems.
\newblock {\em Biological cybernetics}, 52(3):141--152, 1985.

\bibitem[\protect\citeauthoryear{Kaempfer and
  Wolf}{2018}]{kaempfer2018learning}
Yoav Kaempfer and Lior Wolf.
\newblock Learning the multiple traveling salesmen problem with permutation
  invariant pooling networks.
\newblock {\em arXiv preprint arXiv:1803.09621}, 2018.

\bibitem[\protect\citeauthoryear{Khalil \bgroup \em et al.\egroup
  }{2017}]{khalil2017learning}
Elias Khalil, Hanjun Dai, Yuyu Zhang, Bistra Dilkina, and Le~Song.
\newblock Learning combinatorial optimization algorithms over graphs.
\newblock In {\em NIPS}, pages 6348--6358, 2017.

\bibitem[\protect\citeauthoryear{Kobeaga \bgroup \em et al.\egroup
  }{2018}]{kobeaga2018efficient}
Gorka Kobeaga, Mar{\'\i}a Merino, and Jose~A Lozano.
\newblock An efficient evolutionary algorithm for the orienteering problem.
\newblock {\em Computers \& Operations Research}, 90:42--59, 2018.

\bibitem[\protect\citeauthoryear{Kool \bgroup \em et al.\egroup
  }{2018}]{kool2018attention}
Wouter Kool, Herke van Hoof, and Max Welling.
\newblock Attention, learn to solve routing problems!
\newblock {\em arXiv preprint arXiv:1803.08475}, 2018.

\bibitem[\protect\citeauthoryear{Laporte and
  Martello}{1990}]{laporte1990selective}
Gilbert Laporte and Silvano Martello.
\newblock The selective travelling salesman problem.
\newblock {\em Discrete applied mathematics}, 26(2-3):193--207, 1990.

\bibitem[\protect\citeauthoryear{Li \bgroup \em et al.\egroup
  }{2018}]{li2018combinatorial}
Zhuwen Li, Qifeng Chen, and Vladlen Koltun.
\newblock Combinatorial optimization with graph convolutional networks and
  guided tree search.
\newblock In {\em NIPS}, pages 539--548, 2018.

\bibitem[\protect\citeauthoryear{Nazari \bgroup \em et al.\egroup
  }{2018}]{nazari2018reinforcement}
Mohammadreza Nazari, Afshin Oroojlooy, Lawrence Snyder, and Martin Tak{\'a}c.
\newblock Reinforcement learning for solving the vehicle routing problem.
\newblock In {\em NIPS}, pages 9839--9849, 2018.

\bibitem[\protect\citeauthoryear{Nowak \bgroup \em et al.\egroup
  }{2017}]{nowak2017note}
Alex Nowak, Soledad Villar, Afonso~S Bandeira, and Joan Bruna.
\newblock A note on learning algorithms for quadratic assignment with graph
  neural networks.
\newblock {\em stat}, 1050:22, 2017.

\bibitem[\protect\citeauthoryear{Optimization}{2014}]{optimization2014inc}
Gurobi Optimization.
\newblock Inc.,“gurobi optimizer reference manual,” 2015, 2014.

\bibitem[\protect\citeauthoryear{Ramesh \bgroup \em et al.\egroup
  }{1992}]{ramesh1992optimal}
R~Ramesh, Yong-Seok Yoon, and Mark~H Karwan.
\newblock An optimal algorithm for the orienteering tour problem.
\newblock {\em ORSA Journal on Computing}, 4(2):155--165, 1992.

\bibitem[\protect\citeauthoryear{Souffriau \bgroup \em et al.\egroup
  }{2008}]{souffriau2008personalized}
Wouter Souffriau, Pieter Vansteenwegen, Joris Vertommen, Greet~Vanden Berghe,
  and Dirk~Van Oudheusden.
\newblock A personalized tourist trip design algorithm for mobile tourist
  guides.
\newblock {\em Applied Artificial Intelligence}, 22(10):964--985, 2008.

\bibitem[\protect\citeauthoryear{Tang and Miller-Hooks}{2005}]{tang2005tabu}
Hao Tang and Elise Miller-Hooks.
\newblock A tabu search heuristic for the team orienteering problem.
\newblock {\em Computers \& Operations Research}, 32(6):1379--1407, 2005.

\bibitem[\protect\citeauthoryear{Thomadsen and
  Stidsen}{2003}]{thomadsen2003quadratic}
Tommy Thomadsen and Thomas~K Stidsen.
\newblock The quadratic selective travelling salesman problem.
\newblock {\em Informatics and Mathematical Modeling Technical Report}, 2003.

\bibitem[\protect\citeauthoryear{Tsiligirides}{1984}]{tsiligirides1984heuristic}
Theodore Tsiligirides.
\newblock Heuristic methods applied to orienteering.
\newblock {\em Journal of the Operational Research Society}, 35(9):797--809,
  1984.

\bibitem[\protect\citeauthoryear{Van~Hasselt \bgroup \em et al.\egroup
  }{2016}]{van2016deep}
Hado Van~Hasselt, Arthur Guez, and David Silver.
\newblock Deep reinforcement learning with double q-learning.
\newblock In {\em AAAI}, 2016.

\bibitem[\protect\citeauthoryear{Vansteenwegen \bgroup \em et al.\egroup
  }{2011}]{vansteenwegen2011orienteering}
Pieter Vansteenwegen, Wouter Souffriau, and Dirk Van~Oudheusden.
\newblock The orienteering problem: A survey.
\newblock {\em European Journal of Operational Research}, 209(1):1--10, 2011.

\bibitem[\protect\citeauthoryear{Veli{\v{c}}kovi{\'c} \bgroup \em et al.\egroup
  }{2017}]{velivckovic2017graph}
Petar Veli{\v{c}}kovi{\'c}, Guillem Cucurull, Arantxa Casanova, Adriana Romero,
  Pietro Lio, and Yoshua Bengio.
\newblock Graph attention networks.
\newblock {\em arXiv preprint arXiv:1710.10903}, 2017.

\bibitem[\protect\citeauthoryear{Vinyals \bgroup \em et al.\egroup
  }{2015}]{vinyals2015pointer}
Oriol Vinyals, Meire Fortunato, and Navdeep Jaitly.
\newblock Pointer networks.
\newblock In {\em NIPS}, pages 2692--2700, 2015.

\bibitem[\protect\citeauthoryear{Wang \bgroup \em et al.\egroup
  }{1995}]{wang1995using}
Qiwen Wang, Xiaoyun Sun, Bruce~L Golden, and Jiyou Jia.
\newblock Using artificial neural networks to solve the orienteering problem.
\newblock {\em Annals of Operations Research}, 61(1):111--120, 1995.

\end{thebibliography}
